\documentclass[11pt]{article}
\usepackage{pxfonts}
\usepackage[T1]{fontenc}

\usepackage[a4paper]{geometry}
\usepackage{fancyhdr}

\usepackage{hyperref}
\usepackage{latexsym}
\usepackage{url}

\usepackage{textcomp}
\usepackage{dsfont}
\usepackage{graphicx}
\usepackage{booktabs}
\usepackage{multirow}
\usepackage{dirtytalk}

\usepackage{xcolor}
\definecolor{darkblue}{rgb}{0, 0, 0.5}
\hypersetup{colorlinks=true,citecolor=darkblue, linkcolor=darkblue, urlcolor=darkblue}

\usepackage{natbib}
\renewcommand\cite{\citep}	

\newcommand{\WN}[0]{WordNet}

\usepackage{algorithmic}

\usepackage{abstract}

\title{Integrating Approaches to Word Representation}
\author{\href{www.yuvalpinter.com}{Yuval Pinter}
\thanks{\url{uvp@cs.bgu.ac.il}}}
\date{September 2021}

\begin{document}

\newcounter{example}
\newenvironment{example}[1][]{\refstepcounter{example}\par\medskip
    (\theexample) #1 \rmfamily}{\medskip}

\maketitle
\begin{abstract}
    The problem of representing the atomic elements of language in modern neural learning systems is one of the central challenges of the field of natural language processing.
    I present a survey of the distributional, compositional, and relational approaches to addressing this task, and discuss various means of integrating them into systems, with special emphasis on the word level and the out-of-vocabulary phenomenon.
\end{abstract}

\section{Introduction}
\label{intro}

The mission of natural language processing (NLP) as a computational research field is to enable machines to function in human-oriented environments where language is the medium of communication.
We want them to understand our utterances, to connect these utterances with the objects and concepts of the surrounding world, to produce language which is meaningful to us and helps us navigate a task or satisfy an emotional need.
Over the years of its existence, the mainstream of NLP has known shifts motivated by developments in computation, in linguistics, in foundational artificial intelligence, and in learning theory.
Since the mid-2010's, the clear dominant framework for tackling NLP tasks, and an undeniably powerful one, has been that of deep neural networks (DNNs).
This connectionist approach was originally motivated by the workings of the human brain, but has since developed its own characteristics, and formed a well-defined landscape for exploration which includes constraints stemming from the fundamental properties of its design.

This survey focuses on one of these built-in constraints, which I believe to be central to DNNs in the context of natural language, and specifically of text processing, namely that of \textbf{representations}.
DNNs \say{live} in metric space: their operation manipulates real numbers organized into vectors and matrices, propagating function applications and calculated values within instantiations of pre-defined architectures.
This mode of existence is very well-suited to problem domains that inhabit their own metric space, like the physical realms of vision and sound.
In stark contrast to these, the textual form of linguistic communication is built atop a discrete alphabet and hinges on notions such as symbolic semantics, inconsistent compositionality, and the arbitrariness of the sign~\cite{saussure1916cours}.
The example in (\ref{ex:sent}) exhibits all of these: the symbol \textit{dog} refers to two distinct objects bearing no semantic resemblance; \textit{large} and \textit{white} each describe the (canine) dog's physical properties, while \textit{dining} categorizes the table based on its function, and \textit{hot} does not modify (the second) \textit{dog} at all, but rather joins it to denote a distinctive atomic concept.

\vspace{8pt}
\begin{example}
    \label{ex:sent}
    \textit{The large white dog ate the hot dog left on the dining table.}
\end{example}
\vspace{8pt}

Given these properties of language, it is far from straightforward to decide the means by which to transform raw text into an input for a neural NLP system tasked with a goal which requires a grasp on the overall communicative intent of the text, such that this initial representation does not lose basic semantics essential to the eventual outcome.
This transformation process is known as embedding, after which its artifacts are themselves known as \textbf{embeddings}, often used synonymously in context with \say{vectors} or \say{distributed representations}.
Indeed, the choice for default representations has known several shifts within the short DNN era, motivated in part by advances in computational power but also by a collective coming to terms with the limitations of the preceding methods.

The great challenge of representation is compounded by the unboundedness of it all --- human concept space is ever-expanding, and each new concept may be assigned an arbitrary sign (e.g.,~\textit{zoomer}); within an existing concept space, associations capable of inspiring new utterances occupy a combinatorial magnitude which is essentially infinite; and even the form-meaning relationship itself exhibits malleability by humans' interaction with text input devices and various cognitive biases.\footnote{As a case in point, over the course of writing this survey I have manually added dozens of new terms to the Overleaf editor's spell-check dictionary, two in the referring sentence alone.}
Each of these sources of expansion weighs any proposed representational method with the additional burden of generalizing to novel inputs while maintaining consistency in the manner by which they are represented in the system.
In the NLP literature, the surface manifestation of the expanding spaces of concept and form, and of the more locally-constrained disparity between text available at different points in time of a model's training and deployment, is known as the \textbf{out-of-vocabulary} problem, and the unseen surface forms themselves are termed OOVs.

In this survey, I consider three central approaches to representing the fundamental units of natural language text in its input stage and the consequences of each approach's selection on the goals of the systems they are applied in.
The first, most popular, and most successful one when used in isolation, is the \textbf{distributional} approach where the representation function is trained to embed textual units which appear in similar contexts close to each other in vector space.
The second is the \textbf{compositional} approach which seeks to assemble embeddings for workable textual units by breaking them down into more fundamental elements and applying functions over their own representations, less committed to semantic guarantees.
The last is the \textbf{relational} approach which makes use of large semantic structures curated manually or in a semi-supervised fashion, leveraging known connections between text and concepts and among concepts in order to create embeddings manifesting humans' notions of \say{meaning}.
\textbf{The OOV problem} features heavily in the motivation and analysis of the work presented, as it presents challenges to each of the approaches described, yet the exact definition of vocabularies and OOV-ness themselves are challenged by the advent of NLP systems that have become mainstream following the processed described in this work, namely \textbf{contextualized subword embeddings}.

\section{The Atoms of Language}
\label{int:atoms}

Natural language is ultimately a system for conveying meaning, information, and social cues from the realm of human experience into a discrete linear form by encoding them as auditory, visual, and/or textual symbols, which are then iteratively composed into more complex units.
In order to process such a system's outputs by computational means, it seems fitting to identify those symbols which carry the basic units of meaning, and then find the proper ways to map those meanings into representations for a program which can compose them.
The first step, that of identifying linguistic atoms, proves to be a formidable challenge.
From the surface output perspective, the common wisdom is that the basic semantic unit of language is what is known as a \textbf{morpheme}.
The English word \textit{unbelievable}, for example, is composed of a stem morpheme \textit{believe}, a semantic-syntactic suffix \textit{-able} recasting the verb into an adjective pertaining to potential, and a semantic prefix \textit{un-} denoting negation.
But this morpheme = atom stipulation is not unassailable.
Processes below the morpheme level have been documented across languages, for example the sound symbolism phenomenon known as phonaesthesia, where arbitrary sound patterns correlate with a concept or conceptual properties, such as /gl/ in the English light/shine-related words \textit{glow}, \textit{glitter}, and  \textit{glare}~\cite{blake2017sound}.
Less arbitrarily, patterns and even individual sounds in names are known to evoke semantic qualities based on their acoustic properties~\cite{kohler1947gestalt,bergh1984sound}.
In English-language informal communication modes, writers sometimes employ the practice of expressive lengthening, where a single character in a word is repeated in order to amplify its referent's extension on some scale.
For example, \textit{looooong} would be used to describe a particularly long object or period of time.
In addition to these sub-morpheme phenomena, the morpheme symbolism and the atoms of our conceptual space relate at neither a univalent nor a one-to-one relation.
Certain stem morphemes, like \textit{star}, denote multiple types of concepts or objects (\textbf{polysemy} and \textbf{homonymy}), while some concepts may be referred to using different morphemes like the relevant meanings of \textit{room} and \textit{space} (\textbf{synonymy}).
The suffix \textit{-s} can denote both a third-person present verb or a plural noun (\textbf{polyexponence}), and both are replaced by \textit{-es} under certain local conditions (\textbf{flexivity}).

Theoretical quibbles notwithstanding, NLP is a practical field, and from its nascence it was clear that finding the most appropriate way to break text down to its purest elements should not set back our efforts to perform sequence-level tasks and develop useful applications.
Thus, concessions must be made in the form of selecting a unit easily extractable from text and working with it.
This necessity coincides with the reality of having English as the overwhelmingly central target of NLP applications and easiest source of data.
The focus on a language with mostly isolating morphology, where morphemes often occupy distinct word forms that are related through sentence-level syntax, conspired with the technical ease of detecting whitespace in text and led to an inevitable starting point for the community in using the \textbf{space-delimited word} as the basic unit of text analysis.\footnote{I will continue throughout to use \say{space-delimited} to describe a family of simple string tokenization techniques which typically also include minimal heuristics for punctuation separation and a handful of language-specific rules like separating English contractions based on a short closed list, in partial accommodation of the difference between grammatical words and orthographic words~\cite{dixon2002word}.}
The very name of the fundamental bag-of-words approach (BoW) illustrates the implicit synonymity of \say{word} and \say{basic unit of representation} in NLP jargon.
Although subword- and multiword-level systems were designed and developed outside this paradigm, mostly citing a non-English motivation, when the neural revolution came the predominant methods again anchored the field to the space-delimited word as the atom.

The most obvious advantage of this approach is its simplicity, considering how difficult it is in practice to extract correct sub-word morphemes directly from text.
Historically-entrenched orthographic conventions and local-context phonological processes lead to phenomena such as variance in morpheme form at different instantiations, such as the disappearance of the stem's final \textit{e} in the \textit{unbelievable} or the \textit{s}-\textit{t} alteration in derivations like \textit{Mars}-\textit{Martian}, making a deterministic mapping from surface form to morpheme sequence impossible.
The lack of overt textual marking of morpheme boundaries (except for the uncommon case of hyphenation) also leads to ambiguous segmentation in words like \textit{unionize}, and the general property of our sound and writing systems' inventory being relatively small leads to the incidence of affix-identical sequences in single-morpheme words like \textit{reply} (cf. \textit{shortly}) and \textit{bring} (cf. \textit{lying}).
Automatic detection of morphemes can be achieved today by unsupervised data-driven systems like Morfessor~\cite{creutz-lagus-2002-unsupervised,creutz2007unsupervised}, which rely on large amounts of training data and provide no guarantee to finding the true morphemes in all cases or downstream applications.

\section{Neural Representations}
\label{int:reps}

The idea of breaking down concepts in language into numerically-valued axes has played a role in the formation of the modern research landscape in linguistics.
\citet{osgood1952nature} proposed a low-dimensional space in which nominal objects and concepts are represented by values associated with characteristics which may describe them, such that \say{eager} and \say{burning} share a value along the \textit{weak} $\Leftrightarrow$ \textit{strong} dimension, while differing along the \textit{cold} $\Leftrightarrow$ \textit{hot} dimension.
The values were elicited from human subjects.

Scaling this very linguistically-motivated approach manually over an entire language is at the very least impractical, and over the years some relaxations of this scheme to define representations for words which are \textbf{distributed} along dimensions gave rise to more automation-friendly processing techniques.
Most crucial was the realization that the individual dimensions in the representation space do not have to be meaningful in and of themselves.
Liberating the dimensions from their labels allowed the number of dimensions to be governed by concerns of data availability and computational memory and power, rather than by the precision of our semantic theory and ontological thoroughness; it allows for the discovery of unnamed but possibly useful similarities and distinctions between concepts; and it \say{leaves room} for new properties to be learned if, for example, a domain shift occurs during the process of applying an embedding-based system to a downstream task.

Embedding concepts into a \say{blank} vector space using learning methods turns the implied causal direction that motivated Osgood's framework on its head: instead of creating the embeddings based on what we know about language and the relations between concepts, the latter become the proxy target by which we can measure whether or not the embeddings learned by our model are useful to us.
Starting with an arbitrary metric space with well-known properties such as $\mathds{R}^d$ becomes a great advantage, as the space comes with metrics and operations which are easy to conceptualize and imagine as the necessary proxies.\footnote{One heroic departure from the shackles of euclidean space is the line of work on embeddings in hyperbolic space~\cite{nickel2017poincare}, touted as a more suitable representation framework for hierarchical structures, including the semantic structure of a language.}
As the formative instance of this realization served the ability to score the relative directionality of two vectors using the cosine similarity function, which can be compared to annotations in word similarity resources such as WordSim-65~\cite{rubenstein1965contextual}, where human subjects were asked to score word pairs without the hassle of decomposing them into their semantic properties first.
Metric space also affords the intuitive parallelogram metaphor of word analogy, haunting every introductory text and presentation on embeddings with the equation \texttt{king $-$ man $+$ woman $\approx$ queen}.

\section{Distributional Semantics}
\label{int:dist}

The development of the distributed view of representation for linguistic objects accompanied the rise of methodologies making use of the distributional hypothesis, traditionally attributed to~\citet{harris1954distributional} and framed as \say{you shall know a word by the company it keeps}.
The maximalist interpretation of this adage as \say{a word is defined by applying a combination function to the set of its contexts}, used pre-modern-neurally in influential methods such as Brown Clustering~\cite{brown-etal-1992-class}, is an appealing principle to the embedding movement for good reason:
breaking words down into contexts provides us with just the distributed fixed dimensions we seek.
Once we decide exactly what \say{context} means to us, we can
programatically extract all contexts for all target words given only a corpus, and base our latent dimensions (whose number is limited to hundreds or thousands for practical reasons) on them.
The two methods which ended up dominating the distributional embeddings landscape share a definition of context, essentially \say{words that appear near the target word}, but translate this decision into embedding differently.
In SkipGram~\cite{mikolov2013efficient}, dimension significance is built \say{bottom-up} from a random initialization and a traversal of the corpus; in GloVe~\cite{pennington-etal-2014-glove}, dimensions are the result of an implicit reduction of the full $V \times V$ co-occurrence matrix, where $V$ is the number of words in our vocabulary.
The former approach was inspired by early embedding systems~\cite{bengio2003neural} developed around the task of language modeling, which is defined with an expectation based in distributional signals, while the latter has origins in latent semantic analysis~\cite[LSA;][]{deerwester1990indexing}.
Evaluation on intrinsic tasks such as similarity datasets and analogy benchmarks~\cite[e.g.,][]{finkelstein2001placing,mikolov-etal-2013-linguistic,hill-etal-2015-simlex} cemented distributional word embeddings as the representation go-to and an accessible replacement to one-hot encodings for a host of applications, while performance on \textbf{downstream} tasks within deep learning systems advanced the understanding of the utility that \textbf{pre-training} can afford end-to-end systems which include an embedding layer~\cite{collobert2008unified,collobert2011natural}.

\section{Out-of-Vocabulary Words}
\label{int:oov}

The choice of space-delimited words as the basic unit for representation, and the large resource investment necessary to pre-train a distributional model over a large corpus, in both money and time, create a situation where vectors can mostly be trusted \textbf{as long as the words they represent are present in the pre-training corpus}.
The models so far discussed have no intrinsic ability to represent words not present in their lookup table, or out-of-vocabulary, or \textbf{OOV}s~\cite{brill-1995-transformation,brants2000tnt,plank2016non,heigold2017robust,young2018recent}.
Empirical analyses such as the one in \citet{mimick} show that indeed, the overwhelming majority of downstream datasets contain words not present in the pre-training corpora.
\citet{pinter-etal-2020-nytwit} present a diachronical dataset showcasing the volume of novel terms entering a large, steady daily publication in English over time; but even a snapshot of a language at a given moment contains unlimited domain-specific terms, morphological derivations, named entities, potential loanwords, typographical errors, and other sources of OOVs which would appear very reasonably in text analysis tasks and which the downstream model should be given the faculty to handle.
In fact, according to \citet{kornai2002many}, statistical reasoning leads us to conclude that languages have an infinite vocabulary.
But even if a language's word set were finite, and all present in some corpus, practical memory and lookup constraints would still limit embedding tables to non-exhaustive vocabularies.

To overcome the intrinsic limits of corpus-learned embedding tables, the distributional system has begotten some heuristics that try and initialize embeddings for OOVs beyond the trivial random initialization fallback.
If one were to stay true to Firth's maxim, one possible strategy would be to keep SkipGram's context embedding table as well as the main table (for \say{target} words), and initialize OOV embeddings based on the context in which they are first encountered~\cite{horn-2017-context}.
This approach has not caught on, and instead most practitioners took to the use of a special \texttt{<UNK>} embedding, named as an abbreviation of \textit{unknown}~\cite{bengio2003neural}.
In a pre-training stage, such an embedding is learned by replacing a small percentage of the corpus with a dedicated \texttt{<UNK>} token, thus gaining at least some prior for an initialization, in some sense an average over possible contexts for encountering \emph{any} word.
This approach is brutally simplistic;
it assumes not only that all novel words are representable using the same approximation technique, but that they are all \textit{exactly the same}.
The first assumption alone is easy to dispute:
a careful observation of any taxonomy of word formation processes~\cite{lieber2005english,plag2018word} suggests that embedding new words into an existing space must involve considering multiple approaches in parallel.

\begin{itemize}
    \item Words created by processes at the multi-word level, such as compounding or blending, require means of extracting the underlying constructed words and composing the semantic contribution from each word.
    For example, \textit{brunch} is a blend of \textit{breakfast} and \textit{lunch}; a reasonable initial embedding can be the mean vector for these two words, hopefully keeping it at a high similarity with other meals and the appropriate time of day.
    \item Words that are inflections of known words, for example \textit{ameliorating}, can benefit from a morphological analysis which finds its stem and syntactic suffix, placing the new vector at the sum of the verb \textit{ameliorate} and the generalized notion of \textit{-ing} verbs, if one is realized in the embedding space (arguably, in a good space it should at least be reliably extractable).
    \item Novel named entities such as \textit{Lyft} or \textit{SARS-COV-2}, more often than not, reflect arbitrary naming practices and cultural primitives, and even recognition of their type (person / organization / location, etc.) might well be impossible without access to knowledge bases covering the appropriate domain, noting explicitly where in concept space the novel word should be embedded.
    \item Some OOVs are the result of unpredictable subword processes such as typographical errors (typos) and stylistic variation, like the aforementioned expressive lengthening.
    In such cases, it is sometimes best to opt out of creation of a new embedding at all and simply map the new form to the existing embedding of its intended canonical word form.
    This choice will depend on the intended application; in certain cases like sentiment analysis, the stylistic information itself is essential.
    \item Loanwords like \textit{vespa} originate in a different language than the one the embedding was produced for, but in some cases we have access to an embedding space for the origin language and a function which translates between the two languages' space.
    A system which can detect the word and its origin, perhaps overcoming processes like writing-system transliteration and phonological adaptation, can start by embedding the target language word in a position projected from the source language's embedding for the equivalent word form.
\end{itemize}
This is not a comprehensive list.
More types of novel words are identified in \citet{pinter-etal-2020-nytwit}, and not all suggestions in the taxonomy above correspond to actual existing work.
Limiting this discussion to a strict interpretation of written-form uniqueness also prevents us from considering as OOVs concepts which are spelled in the same way as other words, either by chance (homography, for example \textit{row} as a line or a fight), by naming (e.g., \textit{Space Force}), or by processes such as zero-derivation (the verb \textit{smoke}, derived from the noun).
In languages other than English, some OOV-creating forces may be more dominant in word formation than in English.
Morphologically-rich languages, as one edge case, feature large percentages of OOVs in novel texts for a given task's text size compared to English, and this property is often compounded by the fact that many of these are low-resource languages, possessing a relatively small corpus-extracted vocabulary to begin with.

The richness and unpredictability of the OOV problem calls for complementing the word representation systems obtained distributionally with additional approaches, which is the focus of this survey.

\section{Subword Compositionality}
\label{int:comp}

The first approach considered is an attempt to break the space-delimited word paradigm and get at the finer atomic units of meaning, which can then either be used as the fundamental representation layer, or induce better representations at the word level.
This perspective, known as the \textbf{compositional} approach, is inspired mostly by the cases where insufficient generalizations are made for cases of morphological word formation processes.
Under the compositional framework, an ideal representation for \textit{unbelievable} can be obtained by (1) detecting its three morphological components \textit{un-}, \textit{believe}, and \textit{-able}, (2) querying reliable representations learned for each of them, distributionally or otherwise, and (3) properly assembling them via some appropriate function.\footnote{I will use the term \textbf{subword} to denote textual units which are largely between the character level and the word level, when no guarantee of their morphological soundness is attempted. In appropriate contexts, this can also denote word-long or character-long elements which are nevertheless obtained by a subword tokenizer.}

Each of these three steps is a challenge in itself and open to various implementational approaches.
Learning representations for subword units is usually done by considering the subword elements in unison with the full word while applying a distributional method~\cite[e.g.,][]{bojanowski-etal-2017-enriching}, but some have opted for pre-processing the pre-training corpus such that only lemma forms exist as raw text and the other tokens are explicit representations of the morphological attributes attached to each lemma~\cite{avraham-goldberg-2017-interplay,tan-etal-2020-mind}, inducing the production of more consistent vocabularies.
Others yet leave the learning to the downstream task itself, feeding off the backpropagated signal from the training instances~\cite{sutskever2011generating,ling-etal-2015-finding,lample-etal-2016-neural,garneau2019attending}; while others train a compositional network based on the word embedding table in an intermediate phase between pre-training and downstream application~\cite{mimick,zhao-etal-2018-generalizing}.
The composition function from subwords to the word level is also open to many different approaches:
prior work has opted for construction techniques as diverse as using the subword strings as one-hot entries to represent the words themselves~\cite{huang2013learning};
summing morpheme embeddings to produce word embeddings~\cite{botha2014compositional};
traversing a possibly deep morphological parse tree using a recursive neural network~\cite{luong-etal-2013-better};
positing probabilistic word embeddings for which the morpheme embeddings act as a prior distribution~\cite{bhatia-etal-2016-morphological};
side-by-side training of both word-level and character-level modules followed by concatenating the resulting representations, to allow the downstream model to learn from both levels independently and control the interaction terms directly~\cite{plank-etal-2016-multilingual};
assembling a hierarchical recurrent net that progressively encodes longer portions of text in each layer~\cite{chung2019hierarchical};
or dispensing with the word level altogether and just representing text with a single atomic layer of characters or subwords~\cite{sennrich-etal-2016-neural}.

Most challenging of all is the detection of the subwords themselves.
As noted above, morphemes are hard to detect from the surface form of a word.
For the default setting where no curated resources exist to allow correct morpheme extraction from a word's form, as is the case in nearly all languages in the world, the mainstream of compositional representation research has centered on the raw character sequence, the unarguable atom of text,\footnote{At least in languages using the Latin script, like English. Chinese text analysis has benefitted from decomposing characters into strokes or radicals; Hebrew and Arabic include diacritical marks that are not character-intrinsic; and elsewhere, treatment of individual bytes from the Unicode representation of characters has also shown merit.} which is used either via direct operation or as a basis for heuristics that define subword units based on statistical objectives.
The great advantage of using characters or primitive character n-grams as the atomic unit for the model~\cite{santos2014learning,kim2016character,wieting-etal-2016-charagram,bojanowski-etal-2017-enriching,peters-etal-2018-deep} is that it rids us of the need to explicitly designate morphemes altogether; the challenge is to still capture the information they convey, somehow.
In contrast, heuristically learning a subword vocabulary from information-theoretic notions~\cite{sennrich-etal-2016-neural,kudo-richardson-2018-sentencepiece} or character-sequence unigram distribution~\cite{kudo-2018-subword} may find us many true morphemes, but there is no guarantee of either precision or recall: corpus collection effects are significant in determining the ultimate vocabulary, orthographic norms may still obfuscate many useful generalized morphemes, and many frequent character sequences may enter the subword vocabulary as the result of coincidental quirks.
For example, the character sequence \textit{eva} might contribute to the representation of \textit{unbeli\textbf{eva}ble}, passing along signals learned from unrelated words such as \textit{Eva} or \textit{\textbf{eva}luate}.
The ever-growing popularity of systems which use such vocabularies in conjunction with the null composition function that ignores sub-word hierarchy and passes the downstream model embeddings corresponding to the raw subword sequence (see~\S\ref{int:contextual}) prevents any possibility of correcting incorrect subword tokens at the word level: in this scenario, the next processing layer of the model will use the embedding for \textit{eva} as if it were part of the input equally important to a frequent word like \textit{house}.

\section{Relational Semantics}
\label{int:rel}

Another way to complement distributionally-trained embeddings is to incorporate signals from curated type-level \textbf{relational} resources.
The prominent category of such resources is semantic graphs, such as \WN{}~\cite{wordnet} and BabelNet~\cite{navigli-ponzetto-2010-babelnet}, which encode the structural qualities of language as a representation of human knowledge.
The core goal of semantic graphs is to describe connections between referents in the perceived and conceived world, and to this end they make an explicit distinction between words as character sequences and an internal semantic primitive which we can call \textbf{concepts}.
Concepts form the chief node type in the semantic graph, connected by individual edges typed into relations such as hypernymy (\textit{elm} \say{is a} \textit{tree}) or meronymy (\textit{branch} \say{is part of a} \textit{tree}), as well as linguistic facts about concept names (\textit{shop.verb} \say{is derivationally related to} \textit{shop.noun}) which make use of the word-form partition of the graph's node set.
In similar vein, relations which straddle the divide between form and function, like synonymy, are extractable from the bipartite subgraph relating word forms and their available meanings.

In the context of language representation, these structures offer a notion of atomicity stemming from our conceptual primitives, an attractive premise.
They may not answer all needs arising from inflectional morphology (since syntactic properties do not explicitly denote concepts) or some of the other word formation mechanisms, but the rich ontological scaffolding offered by the graph and the prospects of assigning separate embeddings for homonyms in a model-supported manner, assuming sense can be disambiguated in usage, seems much \say{cleaner} than relying on large corpora and heuristics to statistically extract linguistic elements and their meaning.
In addition to this conceptual shift, as it were, the graph structure itself provides a learning signal not present in linear corpus text, relating the basic units to each other through various types of connections and placing all concepts within some quantifiable relation of each other (within each connected component, although lack of any relation path is also a useful signal).
The structure can also occupy the place of the fragile judgment-based word similarity and analogy benchmarks, allowing more exact, refined, well-defined relations to be used for both learning the representations and evaluating them.
Methods which embed nodes and relations from general graph structures before even considering any semantics attached to individual nodes and edges, like Node2vec~\cite{grover2016node2vec} and graph convolutional nets~\cite{gcn}, indeed serve as a basis and inspiration for many of the works in this space.

The fundamentally different manner in which the relational paradigm is complementary to the distributional one in contrast with the compositional one has bearing on the OOV problem, which can be viewed from several perspectives.
First is the potential of semantic graphs to improve representation of words that are rare or not present in a large corpus used to initialize distributional embeddings.
This has proven to be a powerful direction by methods such as retrofitting~\cite{faruqui-etal-2015-retrofitting}, where embeddings of related concepts are pushed together in a post-processing learning phase, showcasing \WN{}'s impressive coverage of English domain-specific taxonomies such as classical natural sciences.
Elsewhere, properly modelling hypernymy, for example, has been found to help understand text with rare words whose hypernyms are well-represented in the pre-training corpus~\cite{shwartz-etal-2017-hypernyms}.\footnote{A tangential but noteworthy approach considers relations that are not curated in large graphs, but rather corpora annotated for inter-word relations such as syntactic dependencies~\cite{madhyastha-etal-2016-mapping}.
Their system creates a mapping between a distributionally-obtained embedding table and one trained on the annotated parses, and generalizes this mapping to words which are now out-of-vocabulary for a further downstream task (e.g., sentiment analysis).
In this case, the reference vocabulary (for defining OOV-ness) is not the unsupervised corpus, but rather an intermediate downstream task.}
Still, semantic graphs provide only a partial solution to the overall goal of OOV impact mitigation, given their limited scope and heavy reliance on expert annotation.

From the other direction, systems relying on semantic graphs for applications such as question answering and dialogue generation are likely to encounter \say{OOVs} of their own, i.e.~words and concepts not present in the underlying graph.
Unlike the corpus-OOV problem, which cannot be quantified convincingly without selecting a specific downstream task first, coping with graph-OOVs can be examined through tasks intrinsic to the graph structure itself.
One such task is \textbf{relation prediction}, where we assume a concept has a known connection with \textit{some} other concept, and need to figure out which one.
Depending on our perspective, either the source or target of the relation may be the OOV concept; for example, on first encounter of the concept \textit{indian lettuce}, we wish to know its hypernym from our set of known concepts.
This task is also useful for a similar class of graphs known as \textbf{knowledge graphs} (KGs), such as Freebase~\cite{bollacker2008freebase}\footnote{Now defunct.} and WikiData~\cite{wikidata}, which differ from semantic graphs in several aspects.
While \WN{} curates connections between semantic concepts and dictionary entries, including certain aspects of the physical world (e.g. \say{an elm is a tree}), KGs focus on real-world entities and often time-sensitive encyclopedic knowledge (e.g. \say{Satya Nadella is the CEO of Microsoft}).
\WN{} is a manually-crafted resource created by language and domain experts, whereas many KGs are either crowdsourced or automatically extracted from databases and large text corpora.
As a result, KGs are typically disconnected, shallow, and sparse, boasting areas of hubness and areas of isolation; this contrasts with semantic graphs, where systematic connectedness and hierarchy have been observed~\cite{sigman2002global}.
KGs are also distinguished by the richness of their relation type variety, in the hundreds or thousands, compared to \WN{}'s 18 relation types (including seven pairs of relations reciprocal to each other).
Nevertheless, much of the work on the relation prediction problem has been developed and evaluated on both semantic and knowledge graphs, as well as on derived tasks like \textbf{graph completion}, where the entirety of a node's connections are to be inferred at once, imitating real-world scenarios of knowledge discovery.

Over the years, distributional methods have been used to feed increasingly complex neural nets predicting relations by embedding both concept nodes and relation edges based on corpus-trained tables, to a large degree of success~\cite[e.g.][]{nickel2011three,socher2013reasoning,bordes2013translating,yang2014embedding,toutanovachen2015,neelakantan2015compositional,ji-etal-2015-knowledge,shi2017proje,dettmers2018conve,nathani-etal-2019-learning}.
The basic idea calls for embedding concepts into a metric space and modeling relations by some operator that induces a score for an embedding pair input, either by translating the concept vectors, combining them via bilinear operators, projecting them onto a \say{scoring scale}, or designing an intricate deep system that finds complex relationships.
While these systems achieve impressive results, they all build on an implicit assumption that relation prediction is a \textbf{strictly local} task: the fit of an edge can be estimated from the nodes it connects and the intended label alone.
In KGs, where structure is of secondary concern, this assumption may go a long way before its limitations stress out performance; in the much more structure-crucial semantic graphs, it is increasingly likely that connections are predicted which should not be permissible from enforceable structural constraints alone, e.g. that the hypernym graph cannot contain cycles.
Some systems indeed go beyond the individual edge to embed and predict relations, for example the idea of a path prediction task~\cite{guu-etal-2015-traversing} which demands more structure reliance, or embedding methods leveraging local neighborhoods of relation interactions and automatic detection of relations from syntactically parsed text in an iterative manner~\cite{riedel-etal-2013-relation,toutanova-etal-2015-representing,gcn}.
Others have constructed prediction models where an adversary produces examples which violate structural constraints such as symmetry and transitivity~\cite{minervini-riedel-2018-adversarially}.
\citet{pinter-eisenstein-2018-predicting} present a system which improves \WN{} prediction by augmenting the distributionally-obtained signal with features (motifs) representing the global structure of the semantic edifice.
In addition to the task benefit, the emerging feature weights lead to discovery of some general properties of English semantics.

\section{Contextualized Representations}
\label{int:contextual}

Recent developments in NLP have brought about a shift in the balance depicted so far with respect to the atomic level chosen to represent language in applications and the approaches taken to create these representations.
Advances in multi-task learning and transfer learning, both in non-neural NLP and in non-NLP deep methods, matured well enough to allow deep NLP to use them effectively as well.
The increase of available computation power and the extreme utility found to lie in recurrent nets, most notably the Long Short-Term Memory cell~\cite[LSTM;][]{hochreiter1997long}, led to a series of works suggesting the incorporation of instance-specific context into the feature extraction part of a model, before applying any task-specific elements, beginning with simple prediction tasks~\cite{melamud-etal-2016-context2vec}, followed by near-full coverage of core NLP~\cite{peters-etal-2018-deep}.
The next step was to continue training the shared-architecture context learner, which we can now safely call a language model, during the downstream step, in a process known as fine-tuning~\cite{howard-ruder-2018-universal}.
Design and processing power considerations, but also downstream performance, fueled the shift~\cite{radford2018improving} from recurrent net infrastructure to transformer models~\cite{vaswani2017attention}, which in turn facilitated another major conceptual innovation where autoregressive token prediction was replaced by masked language modeling, where sequence-medial tokens are hidden from the representation layer and must be predicted based on the remaining context~\cite{devlin-etal-2019-bert,liu2019roberta}.
Throughout this evolution, one main principle remained stable: the language prediction task acts as the pre-training step, providing a scaffolding model which is capable of representing tokens within a sequence at a level of effectiveness that allows downstream tasks to begin training with meaningful \textbf{contextualized} representations.
The heart of contextualization lies in the distributional approach.

The design of these pre-training tasks meant they can no longer tolerate OOV tokens at the rate encountered by static embedding algorithms, as that might render the models unusable for any words that appear in context with OOVs downstream, rather than just the OOVs themselves.
On the other hand, the prediction layer creates a computational bottleneck which scales with the size of the vocabulary, since every token must be available for prediction at all model steps.
Therefore, these models resorted to compositional techniques for the bottom layer where the input sequence is processed into tokens.
The character convolution net selected for ELMo~\cite{peters-etal-2018-deep} did not gain traction, possibly because it didn't provide an adequate method for predicting text from the output layer, and so subsequent models, particularly those relying on transformers, operate over a sequence of equal-status tokens, each representing a word or a subword, from a mid-size vocabulary (tens of thousands) built in a pre-pre-training phase using statistical heuristic techniques mentioned in~\S\ref{int:comp}.
These models inherit the problems endemic to these methods like inadequacy for certain OOV classes, morphological unsoundness, and length-imbalance; as well as issues like the added burden they impose on already limited-length token sequences.
Common wisdom seems to hold that they make up for these shortcomings within the depths of their fully-connected transformer layers, and end up with satisfactory top-layer representations.
Recent work challenging these models with truly novel word forms suggest otherwise~\cite{pinter-etal-2020-nytwit,pinter-etal-2020-will}, while work on either incorporating the compositional signal into subword-vocabulary transformers~\cite{ma-etal-2020-charbert,aguilar2020char2subword,el-boukkouri-etal-2020-characterbert,pinter2021learning}, or replacing the subwords with characters or bytes altogether~\cite{clark2021canine,xue2021byt5}, is rapidly gaining traction as well.

\section*{Acknowledgments}
This survey is an adapted version of the introduction my PhD thesis.
I thank my committee for helping to shape it: my advisor, Jacob Eisenstein; Mark Riedl, Dan Roth, Wei Xu, and Diyi Yang.

\bibliography{references,anthology,cite-strings,cite-definitions}

\end{document}